\documentclass[10pt,twocolumn,letterpaper]{article}

\usepackage[pagenumbers]{iccv} 

\definecolor{iccvblue}{rgb}{0.21,0.49,0.74}
\usepackage[pagebackref,breaklinks,colorlinks,allcolors=iccvblue]{hyperref}

\def\paperID{10918} 
\def\confName{ICCV}
\def\confYear{2025}

\title{Unleashing the Multi-View Fusion Potential: Noise Correction in VLM for Open-Vocabulary 3D Scene Understanding}

\author{Xingyilang Yin$^1$\footnotemark[1]~, Jiale Wang$^1$\footnotemark[1]~, Xi Yang$^1$\footnotemark[2]~, Mutian Xu$^2$, Xu Gu$^1$, Nannan Wang$^{1}$ \\
\small{$^1$Xidian University} \small{$^2$SSE, CUHKSZ}\\
}

\begin{document}
\maketitle

\begin{abstract}
Recent open-vocabulary 3D scene understanding approaches mainly focus on training 3D networks through contrastive learning with point-text pairs or by distilling 2D features into 3D models via point-pixel alignment. While these methods show considerable performance in benchmarks with limited vocabularies, they struggle to handle diverse object categories as the limited amount of 3D data upbound training strong open-vocabulary 3d models. We observe that 2D multi-view fusion methods take precedence in understanding diverse concepts in 3D scenes. However, inherent noises in vision-language models lead multi-view fusion to sub-optimal performance. To this end, we introduce \textbf{MVOV3D}, a novel approach aimed at unleashing the potential of 2D \textbf{m}ulti-\textbf{v}iew fusion for \textbf{o}pen-\textbf{v}ocabulary \textbf{3D} scene understanding. We focus on reducing the inherent noises without training, thereby preserving the generalizability while enhancing open-world capabilities. Specifically, MVOV3D improves multi-view 2D features by leveraging precise region-level image features and text features encoded by CLIP encoders and incorporates 3D geometric priors to optimize multi-view fusion. Extensive experiments on various datasets demonstrate the effectiveness of our method. Notably, our MVOV3D achieves a new record with 14.7\% mIoU on ScanNet200 and 16.2\% mIoU on Matterport160 for challenge open-vocabulary semantic segmentation, outperforming current leading trained 3D networks by a significant margin. 

\end{abstract}    
\vspace{-0.5cm}
\section{Introduction}
\label{sec:intro}

\begin{figure}[t]
  \centering
   \includegraphics[width=0.98\linewidth]{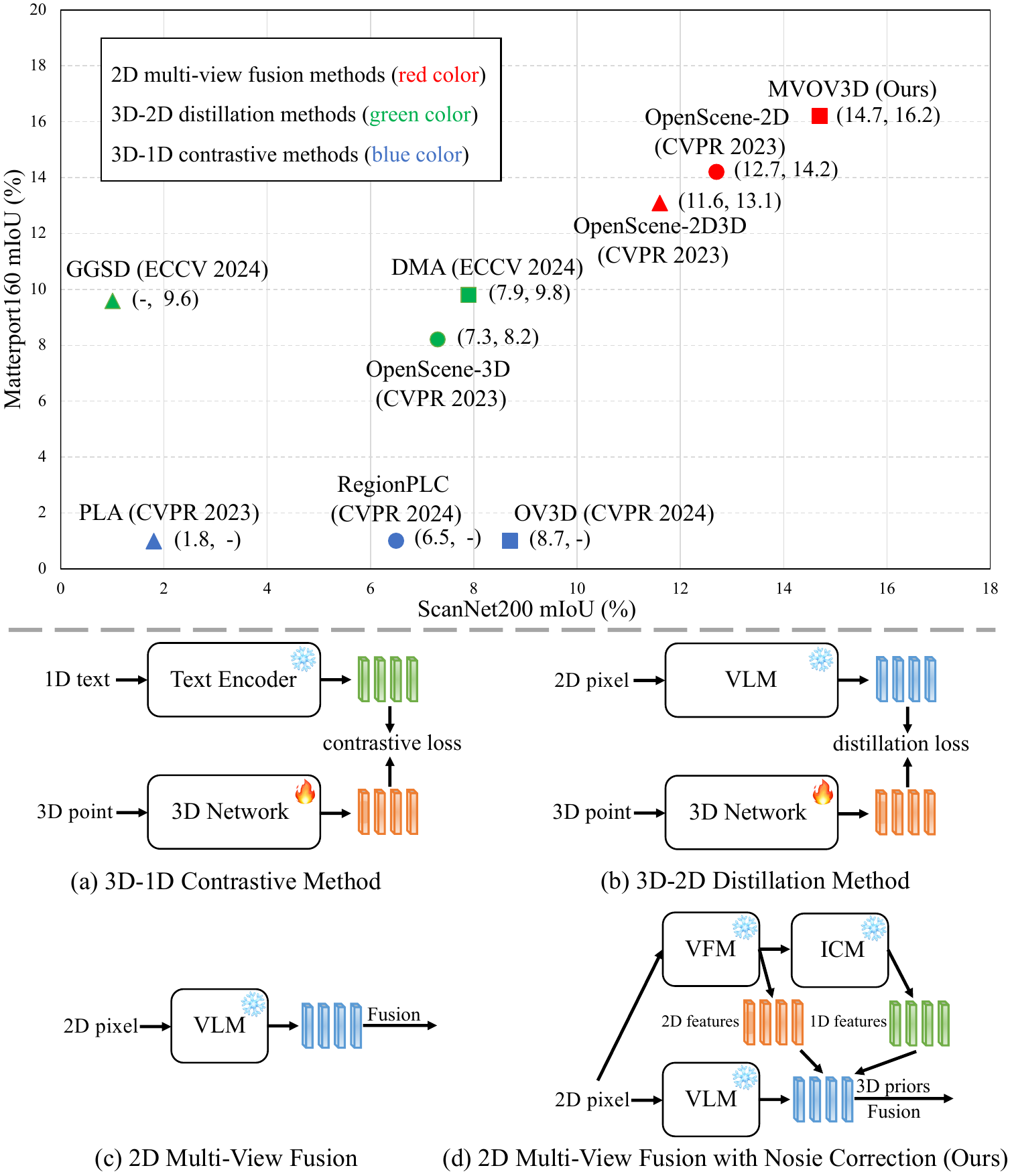}
   \vspace{-0.1cm}
   \caption{\textbf{Top:} Performance comparison on ScanNet200 and Matterport160 for challenging open-vocabulary 3d semantic segmentation. \textbf{2D multi-view fusion} methods take precedence over 3D-2D distillation frameworks and 3D-1D contrastive approaches. The proposed MVOV3D achieves a new record with 14.7\% mIoU on ScanNet200 and 16.2\% on Matterport160. \textbf{Down:} Illustration of different open-vocabulary 3D scene understanding methods.}
   \vspace{-0.6cm}
   \label{fig:figure1}
\end{figure}

3D scene understanding is a fundamental perception task with numerous real-world applications including robot manipulation, virtual reality, and autonomous driving. Traditional close-set methods are trained on datasets with fixed and known categories~\cite{qi2017pointnet++, zhao2021point, misra2021end, wang2024uni3detr, wu2024point}, impeding their potential applications in dynamic and previously unseen 3D environments. This motivates researchers to develop open-vocabulary approaches for 3D scene understanding in real-world settings, where the possible categories are neither predefined nor limited.

Foundation vision-language models (VLM)~\cite{radford2021learning, jia2021scaling}, trained on internet-scale paired image and text data, have demonstrated impressive zero-shot recognition capabilities through a co-embed visual and language feature space. 2D open-vocabulary works have made significant success in various image understanding tasks such as open-vocabulary detection~\cite{li2022grounded, zhou2022detecting, cheng2024yolo} and segmentation~\cite{lilanguage, ghiasi2022scaling, liang2023open, yu2023convolutions}, by leveraging pre-trained VLM. However, compared to the available billions of image and text data, 3D data is relatively scarce. Thus, directly applying pre-training on large-scale 3D-text pairs is not feasible in the 3D domain.

To address it, a straightforward strategy (\textit{3D-1D contrastive}) is to train 3D networks to compare 3D features with text features~\cite{ding2023pla, yang2024regionplc, jiang2024open}  on point-text pairs, as shown in Fig.~\ref{fig:figure1} (a). For instance, PLA~\cite{ding2023pla} and RegionPLC~\cite{yang2024regionplc} employ image caption models~\cite{wang2022ofa, peng2023kosmos, zhou2022detecting} to generate textual descriptions for multi-view images, which are then associated with 3D points. This enables 3D open-vocabulary scene understanding by directly training 3D models through point-text contrastive learning. OV3D~\cite{jiang2024open} further enhances this by establishing fine-grained point-text correspondence through point-to-EntityText alignment. Fig.~\ref{fig:figure1} (b) illustrates another typical strategy (\textit{3D-2D distillation}) in recent works~\cite{peng2023openscene, li2024dense, wang2024open} involving training 3D networks to align point features with image features, which are already associated with text features extracted from pixel-aligned VLM~\cite{ghiasi2022scaling, lilanguage, yu2023convolutions}. For example, with the alignment of the point and its corresponding 2D multi-view pixels, OpenScene-3D~\cite{peng2023openscene} transfers 2D open-vocabulary capabilities to the 3D network via distillation. GGSD~\cite{wang2024open} and DMA~\cite{li2024dense} improve the distillation process through geometry guidance and dense multimodal relations, respectively. While the previous works~\cite{peng2023openscene, wang2024open, li2024dense} demonstrate the priority of training 3D networks over 2D multi-view fusion on small-vocabulary datasets such as ScanNet~\cite{dai2017scannet} with 20 semantic labels and Matterport3D~\cite{chang2017matterport3d} with 21 common categories, they struggle with more diverse object classes.  

As shown in Fig.~\ref{fig:figure1}-Top, under more challenging open-vocabulary settings on ScanNet200~\cite{rozenberszki2022language} and Matterport160 (Matterport3D with 160 semantic labels), training 3D models exhibits poor performance, whether applying explicit point-text contrastive learning or implicit point-pixel distillation. These recently proposed methods perform even worse than OpenScene-2D~\cite{peng2023openscene}, which represents the 3D scene as 2D multi-view images and fuses the 2D multi-view features into corresponding 3D point features (as shown in Fig.~\ref{fig:figure1} (c)). Moreover, OpenScene-2D3D, ensembled features from both the 2D and 3D domains, fails to boost the performance. These observations reveal that although current 3D-2D distillation and 3D-1D contrastive methods leverage 2D foundation models and try to transfer the open-vocabulary capabilities to 3D, the learned 3D features cannot understand diverse concepts in real-world 3D scenes. \textbf{Training 3D networks on a relatively small 3D dataset} limits their open-vocabulary 3D understanding ability, as 3D datasets are relatively small for training a strong 3D network with unbounded vocabularies.

To this end, we abandon the 3D networks and propose MVOV3D to unleash the potential of 2D multi-view fusion for open vocabulary 3D scene understanding, as illustrated in Fig.~\ref{fig:figure1} (d). We aim to \textbf{preserve the open-vocabulary capabilities of pre-trained VLMs} and focus on \textbf{alleviating inherent noises in these models without training}. We refine the 2D multi-view features from 1D, 2D, and 3D perspectives. Specifically, we observe that the incorrect predictions in 3D point features often stem from noise in 2D features. VLM tends to perform poorly when facing occlusion and viewpoint changes. To address this, we leverage vision foundation models (VFM)~\cite{kirillov2023segment, liu2023grounding} to isolate individual entities with high confidence and apply the CLIP \cite{radford2021learning} vision encoder to encode each image region independently. 2D features can be refined by the features extracted through the improved image regions. Additionally, we utilize image caption models (ICM)~\cite{lidecap, huang2023open, li2022blip, zhang2024recognize} to generate high-confidence text descriptions for these regions, further improving the 2D features using the CLIP text encoder. Finally, we assume that points with similar geometry in 3D space should exhibit similar features, thus we guide 2D multi-view fusion with geometric priors by pooling the point features within the same superpoints.

Our main contributions are summarized as follows:
\begin{itemize}[itemsep=0.1pt,topsep=3pt,leftmargin=*]
    \item We analyze the limitations of point-text contrastive learning and point-pixel distillation methods, which transfer the open-vocabulary capabilities to the 3D domain by training a 3D network on relatively small 3D datasets. Instead, we propose preserving the open-vocabulary capabilities of pre-trained VLM without training.
    \item We focus on alleviating inherent noise in VLM. We propose a novel framework that leverages several foundation models (e.g., VFM to extract better image region, ICM to obtain precise text descriptions) and 3D geometric priors to refine the 2D multi-view features from 2D, 1D, and 3D perspectives, enabling improved understanding of diverse object categories in 3D scenes.
    \item We conduct extensive experiments under challenging open-vocabulary settings, including ScanNet200~\cite{rozenberszki2022language}, Matterport3D~\cite{chang2017matterport3d}, and Replica~\cite{straub2019replica} datasets. The results demonstrate our MVOV3D sets a new state-of-the-art for open-vocabulary 3D scene understanding.
\end{itemize}
\vspace{-0.1cm}
\section{Related Work}
\label{sec:related_work}

\paragraph{Closed-Set 3D Scene Understanding.} Previous works have made significant progress on closed-set 3D understanding tasks including 3D semantic segmentation~\citep{qi2017pointnet++, choy20194d, zhao2021point, xu2021paconv, qian2022pointnext, yin2024point, wu2024point}, 3D instance segmentation~\citep{jiang2020pointgroup, schult2023mask3d, ngo2023isbnet, jain2024odin}, and 3D object detection~\citep{shi2019pointrcnn, zheng2021se, wang2022detr3d, wang2024uni3detr}. However, they are all typically trained on predefined categories, which limits their application in real-world scenarios.

\begin{figure*}[t]
  \centering
   \includegraphics[width=0.88\linewidth]{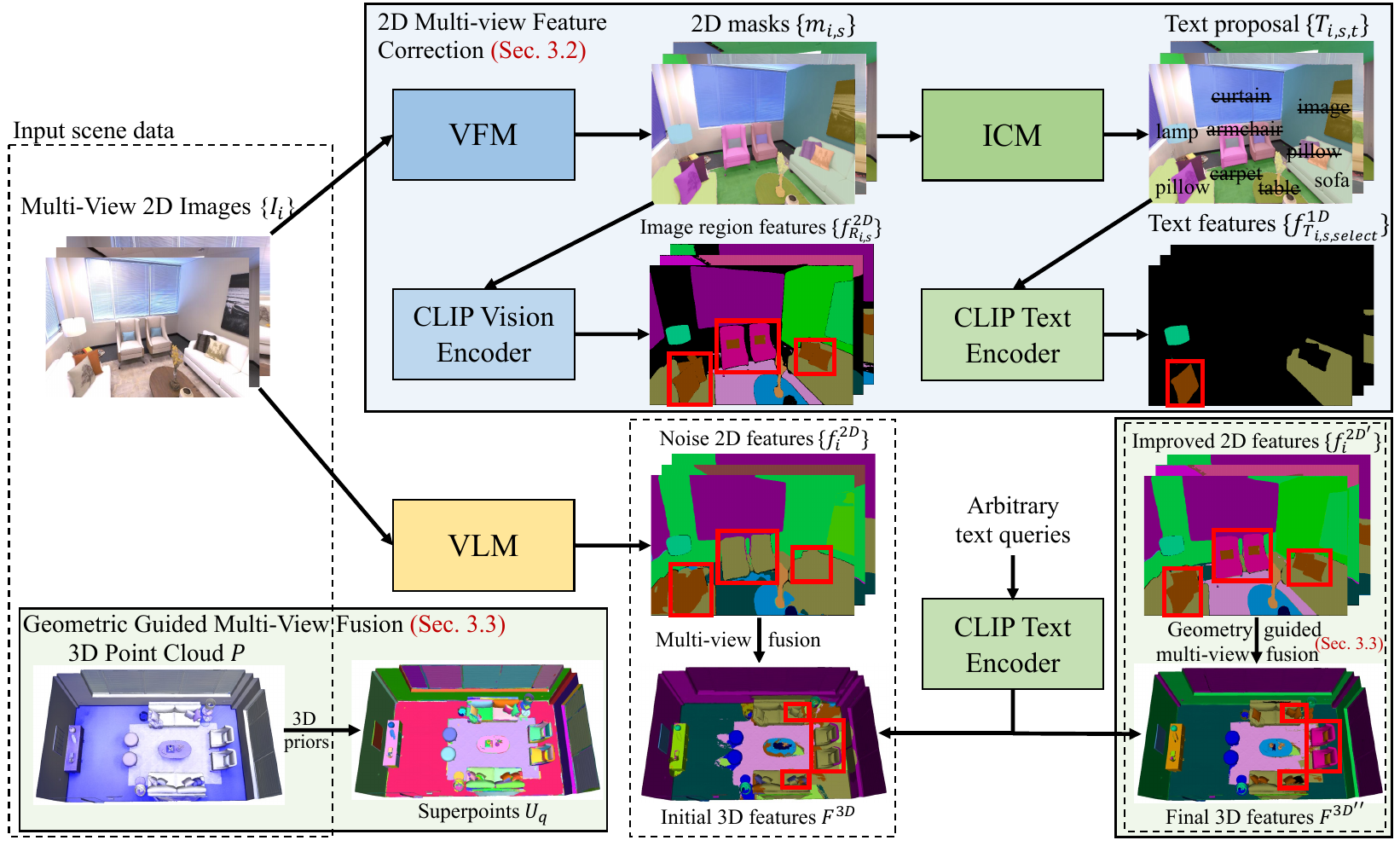}
   \vspace{-0.2cm}
   \caption{Overview of our MVOV3D framework. MVOV3D takes a 3D point cloud $P$ and a set of $N$ corresponding multi-view 2D images $\{I_i\}_{i=1,...,N}$ as input, and corrects the inherent noises in 2D multi-view features $\{f^{2D}_{i}\}$ from 1D, 2D and 3D perspectives. To alleviate the noise, precise 2D masks $\{m_{i,s}\}$ are obtained by vision foundation models (VFM) to isolate image regions for the vision-language model to extract 2D image region features $\{f^{2D}_{R_{i,s}}\}$, and vocabulary-rich image caption models (ICM) are employed to generate precise text descriptions to obtain 1D text features $\{f^{1D}_{T_{i,s,select}}\}$. Finally, 3D geometric priors are applied to guide the multi-view fusion process.}
   \vspace{-0.4cm}
   \label{fig:figure2_overview}
\end{figure*}

\vspace{-0.5cm}
\paragraph{Open-Vocabulary 2D Scene Understanding.} The community has increasingly concentrated on understanding objects beyond a closed set of categories, aiming to address open-world challenges. Milestone VLMs~\citep{radford2021learning, jia2021scaling}, trained on huge amounts of text-image pairs, demonstrate remarkable zero-shot capabilities in 2D vision. Recent studies further explore fine-grained relations, such as pixel-level alignment between visual and language features \citep{lilanguage, ghiasi2022scaling, liang2023open, xu2023open, yu2023convolutions}. In this paper, we leverage pre-trained VLM to extract 2D multi-view features. Further, we alleviate inherent noise in VLM from 1D, 2D and 3D spaces, using VFMs~\cite{kirillov2023segment, liu2023grounding}, ICMs~\cite{lidecap, huang2023open, li2022blip, zhang2024recognize}, and 3D geometric priors.

\vspace{-0.6cm}
\paragraph{Open-Vocabulary 3D Scene Understanding.} Recent methods of open-vocabulary 3D scene understanding can be classified into 4 groups. The first group~\citep{takmaz2023openmask3d, nguyen2024open3dis, yang2023sam3d, xu2023sampro3d, yin2024sai3d, guo2024sam, yang2024sa3dip, boudjoghra2024open} focuses on generating class-agnostic 3D instances from 3D instance proposal networks or back-projected 3D masks from 2D instance masks. Their open features need to be further extracted by either point-wise feature extraction~\citep{takmaz2023openmask3d, yin2024sai3d} or mask-wise feature extraction ~\citep{nguyen2024open3dis}. The second group includes PLA~\citep{ding2023pla}, RegionPLC~\citep{yang2024regionplc}, and OV3D~\citep{jiang2024open}, which explicitly aligns point cloud features with open text features through contrastive learning. The third group~\citep{chen2023clip2scene, peng2023openscene, li2024dense, wang2024open, zhu2024open} distills 2D pixel features to 3D points using point-pixel alignment. These approaches implicitly achieve open-vocabulary 3D understanding, as the 2D pixel features are derived from pixel-aligned 2D open-vocabulary models~\citep{lilanguage, ghiasi2022scaling}. However, methods in the former groups exhibit considerable performance in identifying a limited number of categories yet degrade their ability when confronted with a wide variety of classes. The fourth group comprises ConceptFusion~\citep{jatavallabhula2023conceptfusion}, OVIR-3D~\cite{lu2023ovir}, and OpenScene~\citep{peng2023openscene}, which aggregates 2D multi-view features onto 3D points. Following the final group, we propose MVOV3D to mitigate incorrect predictions in 2D VLM without training, thereby maximizing the open-vocabulary capabilities and achieving state-of-the-art performance in open-vocabulary 3D scene understanding. 

\section{Method}
\label{sec:method}
In this work, we propose MVOV3D, a 2D multi-view fusion-based framework for open-vocabulary 3D scene understanding. Our objective is to preserve the open-vocabulary capabilities of pre-trained VLMs and address inherent noises in these models without training. We refine the 2D multi-view features from 1D, 2D, and 3D perspectives with several foundation models.

\begin{figure*}[t]
  \centering
   \includegraphics[width=0.84\linewidth]{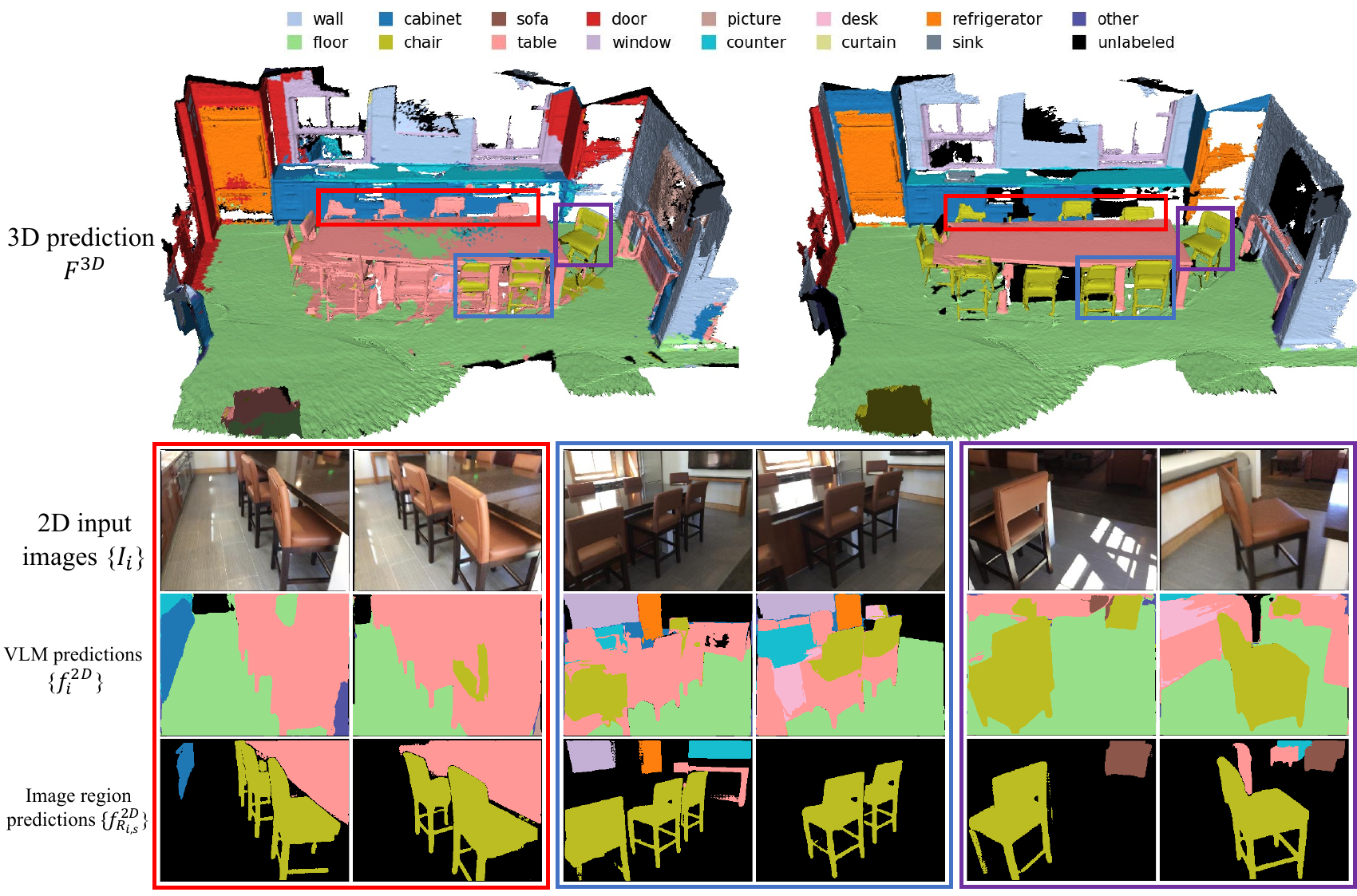}
   \vspace{-0.2cm}
   \caption{Illustration of 3D predictions and corresponding 2D predictions generated by VLM. Inaccurate 3D predictions arise from noise in the 2D predictions (highlighted in the \textcolor{red}{red} and \textcolor{blue}{blue} boxes). The VLM struggles with co-occurrence issues but performs better when processing isolated entities (highlighted in the \textcolor{purple}{purple} box). To address this, we refine the 2D multi-view features from 2D space by segmenting the 2D images into distinct image regions using vision foundation models.}
   \vspace{-0.3cm}
   \label{fig:figure3_2d_refine}
\end{figure*}

\subsection{Overview}
\label{sec3.1:overview}
An overall architecture of MVOV3D is illustrated in Fig.~\ref{fig:figure2_overview}. For each scene, the framework takes a 3D point cloud $P$ and a set of $N$ corresponding multi-view 2D images $\{I_i\}_{i=1,...,N}$ as input, and  outputs per-point features for open-vocabulary queries. The initial step involves extracting pixel features for images through pixel-aligned VLMs~\cite{ghiasi2022scaling, lilanguage}. Due to the inherent noise, these VLMs underperform when handling occlusion and viewpoint changes within images, leading to inaccurate 3D point features after multi-view fusion. To enhance feature quality, we refine the pixel features from 1D, 2D, and 3D perspectives using various foundational models. We address the noise issues by employing 2D VFMs~\cite{kirillov2023segment, liu2023grounding} to obtain image regions with high confidence. The corresponding pixel features can be improved by fusing these better 2D image regions encoded by the CLIP vision encoder. Subsequently, we utilize image caption models~\cite{lidecap, huang2023open, li2022blip} to generate text descriptions and extract object names with high confidence for each image region, leveraging the extensive vocabularies in these models. These 1D text descriptions further improve the 2D pixel features within a shared image-language feature space via the CLIP text encoder. Finally, assuming that points with similar 3D geometry should exhibit similar features, we guide the multi-view fusion process using 3D geometric priors by pooling the point features within the same superpoints.



\subsection{2D Multi-View Features Correction}
\label{sec3.2:refine2D_1D}
Given a set of $N$ multi-view 2D RGB images $\{I_i\}_{i=1,...,N}$ with resolution of $H\times W$, we utilize a pre-trained pixel-aligned VLM~\cite{ghiasi2022scaling, lilanguage} to extract dense per-pixel features $F^{2D}=\{f^{2D}_{1}, ..., f^{2D}_{N}\}$, where $f^{2D}_{i} \in\mathbb{R}^{H \times W \times C}$, and $C$ is the dimension of open-vocabulary features. For a scene $P$ containing $M$ points, the 3D point features $F^{3D}\in\mathbb{R}^{M \times C}$ can be simply obtained by applying a multi-view fusion strategy. Specifically, average pooling is performed over the corresponding 2D pixel features for each point via 3D-2D alignment~\cite{peng2023openscene, wang2024open}. 

\begin{figure}[t]
  \centering
   \includegraphics[width=0.88\linewidth]{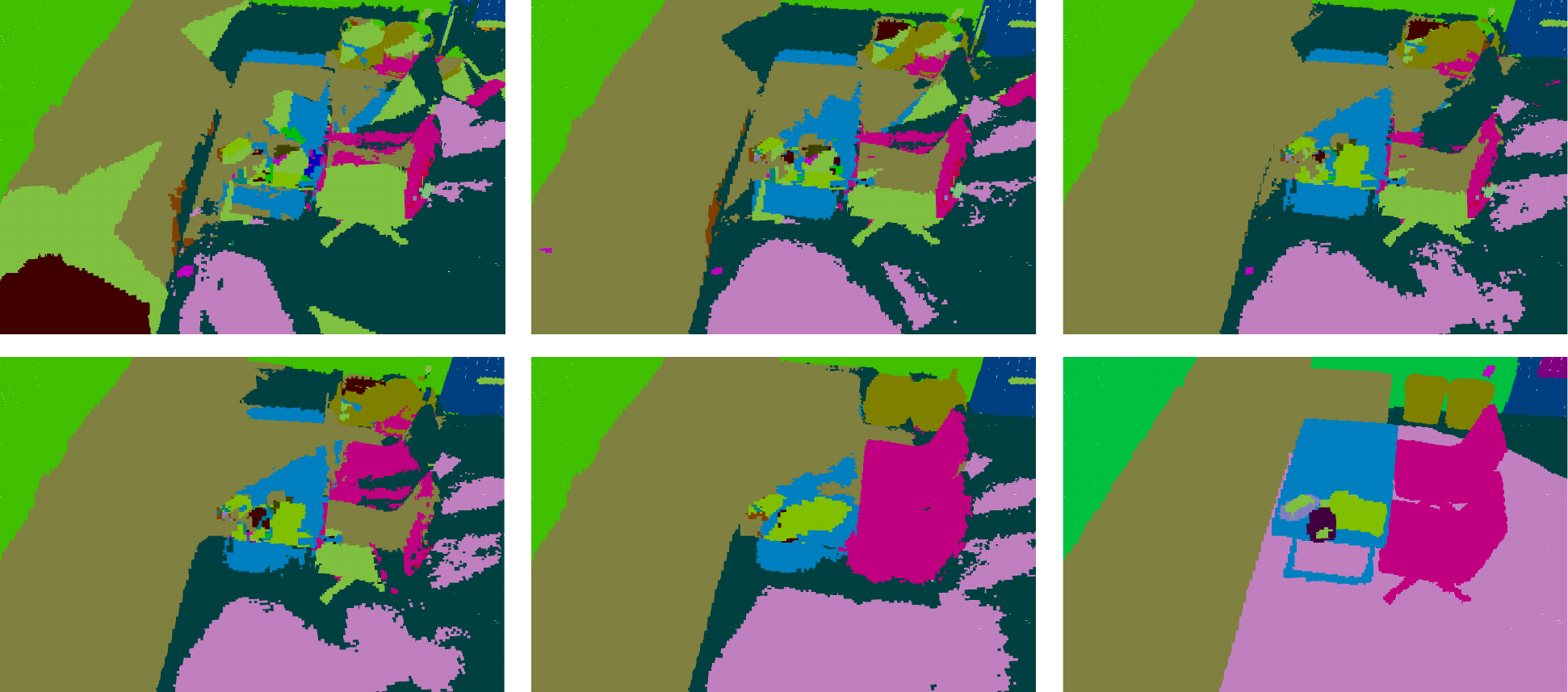}
   \vspace{-0.1cm}
   \caption{Illustration under different occlusion threshold on Replica. Top: 0.7, 0.6, 0.5; Bottom: 0.4, 0.0, GT.}
   \vspace{-0.2cm}
   \label{fig: occlusion}
\end{figure}

However, the 3D features $F^{3D}$ often yield inaccurate predictions and exhibit variability even within the same object class. As shown in the 3D predictions of Fig.\ref{fig:figure3_2d_refine}, the chairs in the \textcolor{red}{red} box are completely misclassified, the chairs in the \textcolor{blue}{blue} box contain partial errors, while the chair in the \textcolor{purple}{purple} box is correctly identified. To investigate this issue, we illustrate the predictions of the VLM on the 2D images before the multi-view fusion. As shown in the second row of images in Fig. \ref{fig:figure3_2d_refine}, the chairs are misclassified as tables when chairs and tables co-occur, but are correctly identified when the chair instance is processed without occlusions. This observation indicates that the 3D features often produce incorrect predictions due to errors in the 2D multi-view features, and VLM tends to struggle with occlusions. Additional quantitative results under occlusion are shown in Fig. \ref{fig: occlusion} and Table \ref{tab: occlusion}. We evaluate the performance of VLM like OpenSeg~\cite{ghiasi2022scaling} under various occlusion settings on Replica dataset. Specifically, when the occlusion threshold is set to 0.4, this means if fewer than 40\% of a 3D instance’s points are valid in a given image, we use its 2D features for multi-view fusion. The results show that as the level of \textbf{\textit{occlusion increases}}, the quality of the features extracted by OpenSeg \textbf{\textit{degrades}} accordingly.

\begin{table}[t]
	\centering
	\begin{tabular}{c|c|c|c|c|c}
         \hline
		 Occlusion thres. & 0.0 & 0.4 & 0.5 & 0.6 &0.7 \\
	\hline
         mIoU (\%) &\textbf{18.2} & 16.5 &14.8 & 13.5 & 11.7 \\
        \hline
	\end{tabular}
        \vspace{-0.2cm}
        \caption{Quantitative occlusion results on Replica.}
        \vspace{-0.5cm}
	\label{tab: occlusion}
\end{table}


To address this co-occurrence issue, we propose an intuitive approach that segments the images to isolate individual entities and apply the VLM to extract features from each image region independently. Specifically, we utilize pre-trained 2D vision foundation models (VFM), such as Grounding-DINO~\cite{liu2023grounding} and SAM~\cite{kirillov2023segment} to exploit their universal class-agnostic segmentation capabilities. These models segment each image into a set of 2D masks $m_{i,s} \in\mathbb{R}^{H\times W}$ with high confidence. These masks are used to isolate $S$ image regions ${\{R_{i,s}\}}_{s=1, ..., S}$ for image $i$, and the features $f_{R_{i,s}}\in\mathbb{R}^{C}$ for each region are then extracted using the CLIP vision encoder. The $i$th image region features, denoted as $f^{2D}_{R_{i,s}}\in \mathbb{R}^{H \times W \times C}$ are computed by propagating the features across each 2D mask:
\vspace{-0.15cm}
\begin{equation}
\vspace{-0.15cm}
    f^{2D}_{R_{i,s}}(h, w, c) = m_{i,s}(h, w) \cdot f_{R_{i,s}}(c),
    \label{eq:1}
\end{equation}
where $f^{2D}_{R_{i,s}}(h, w, c)$ is the resulting 2D feature map at position $(h, w)$ with feature channel $c$, $m_{i,s}(h, w)=\{0, 1\}$ indicates whether the pixel at $(h, w)$ belongs to $R_{i,s}$, and $f_{R_{i,s}}(c)$ is the value of the CLIP vision feature at channel $c$. Thus, in this process, if $m_{i,s}(h, w) = 1$, then $f^{2D}_{R_{i,s}}(h, w, c) = f_{R_{i,s}}(c)$, otherwise $f^{2D}_{R_{i,s}}(h, w, c) = 0$.

As illustrated in the final row of Fig.~\ref{fig:figure3_2d_refine}, the predictions for image regions generated by VFM and encoded by CLIP vision encoder are more accurate than the original VLM predictions when handling co-occurrence cases, which improves 2D multi-view features from 2D perspectives. 

\begin{figure}[t]
  \centering
   \includegraphics[width=0.8\linewidth]{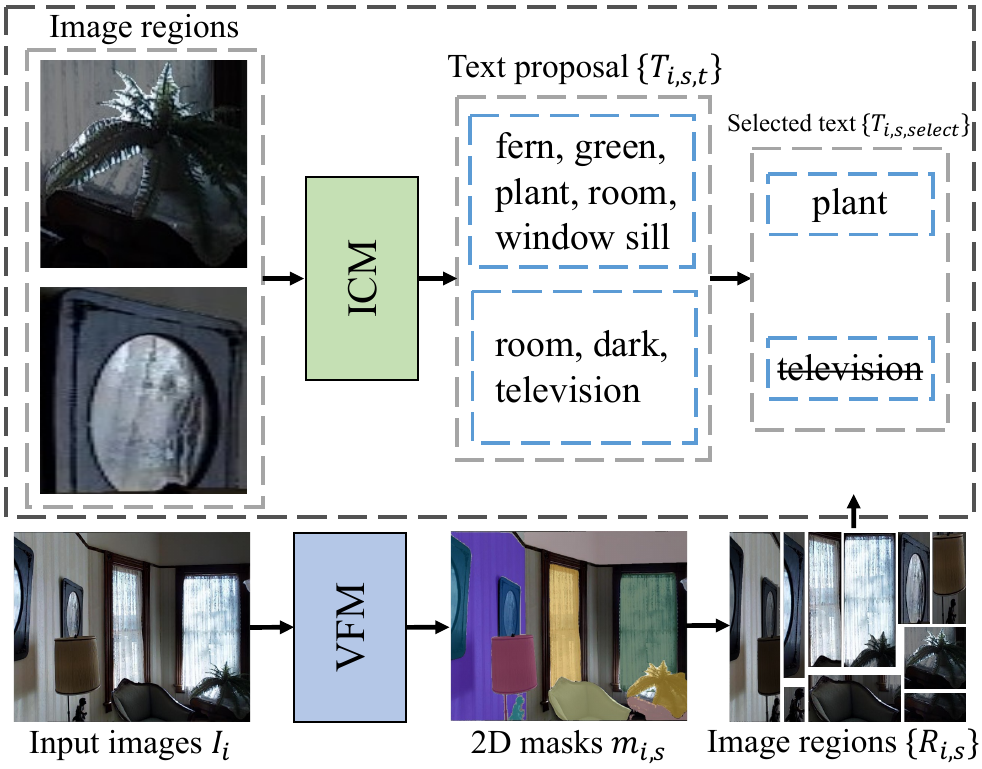}
   \vspace{-0.1cm}
   \caption{An example of correcting the noise in 2D multi-view features with image caption models.}
   \vspace{-0.3cm}
   \label{fig:figure4_1d_refine}
\end{figure}

Subsequently, we further refine 2D multi-view features by leveraging comprehensive vocabularies provided by image caption models (ICM)~\cite{lidecap, zhang2024recognize, huang2023open, li2022blip}. Fig. \ref{fig:figure4_1d_refine} shows an example of the 1D refinement process. We feed image regions generated by VFM into image caption models to obtain relevant text descriptions. For ICM such as Decap~\cite{lidecap} and BLIP~\cite{li2022blip}, which generate a detailed sentence like ``there is a chair and a lamp in a room with a clock on the wall and a window behind it and a lamp on the floor in front of the window'', we use NLP tool like SpaCy~\cite{honnibal2017spacy} to extract $T$ noun phrases like [chair, lamp, room, clock, wall, window, floor, window]. For models like RAM~\cite{zhang2024recognize} and RAM++~\cite{huang2023open}, we directly use the output tags as text proposals, denoted as ${\{T_{i,s,t}\}}_{t=1, ..., T}$. The next step is to assign appropriate text to each image region. To achieve it, we employ CLIP text encoder to extract text embeddings $f_{T_{i,s,t}}\in\mathbb{R}^{C}$ for each text proposal and calculate the cosine similarity scores between $f_{T_{i,s,t}}$ and $f_{R_{i,s}}$. Based on these similarity scores, we specify the text with the highest correspondence. However, in some cases, the selected text may still be inaccurate, such as selecting ``television'' for a mirror region, as shown in Fig. \ref{fig:figure4_1d_refine}. To address it, we apply a confidence threshold $\delta$ to filter out selected text with low confidence. Formally, the 1D refinement process can be defined as:
\vspace{-0.2cm}
\begin{equation}
\vspace{-0.2cm}
    T_{i,s,select} = \arg\max_{t}(\cos(f_{T_{i,s,t}}, f_{R_{i,s}})>\delta),
    \label{eq:2}
\end{equation}
where $T_{i,s,select}$ is the selected text description and the corresponding left masks turn to be  $m_{i,s,select}$. The text features $f_{T_{i,s,select}}\in\mathbb{R}^{C}$ for region $R_{i,s}$ can be extracted by CLIP text encoder. Finally, the 1D text features $f^{1D}_{T_{i,s,select}}\in \mathbb{R}^{H \times W \times C}$ can be calculated by flood-filling each 2D mask with the corresponding text embeddings:

\vspace{-0.3cm}
\begin{equation}
    f^{1D}_{T_{i,s,select}}(h, w, c) = m_{i,s,select}(h, w) \cdot f_{T_{i,s,select}}(c).
    \label{eq:3}
\end{equation}

Once we have the 2D per-pixel features $f_{i}^{2D}$, 2D image regions features $f_{R_{i,s}}^{2D}$, and 1D text features $f^{1D}_{T_{i,s,select}}$, the improved 2D per-pixel features $f_{i}^{{2D}^{'}}$ can be computed as:


\vspace{-0.35cm}
\begin{equation}
    \begin{split}
    f_{i}^{{2D}^{'}}(h, w) = (f_{i}^{2D}(h, w) + f_{R_{i,s}}^{2D}(h, w) + \\
    f^{1D}_{T_{i,s,select}}(h, w)) /  (\mathbb{I}({f_{i}^{2D}(h, w)})+ \\ \mathbb{I}({f_{R_{i,s}}^{2D}(h, w)}) + \mathbb{I}(f^{1D}_{T_{i,s,select}}(h, w))),
    \label{eq:4}
    \end{split}
\end{equation}
 where $\mathbb{I}(x)$ = \{0, 1\} is the indicator function. If $x = 0$, then $\mathbb{I}(x) = 0$; otherwise, $\mathbb{I}(x) = 1$. The improved 2D per-pixel features $f_{i}^{{2D}^{'}}$ are computed by averaging the valid values of the 2D per-pixel features $f_{i}^{2D}$, 2D image regions features $f_{R_{i,s}}^{2D}$, and 1D text features $f^{1D}_{T_{i,s,select}}$ at pixel $(h, w)$. The 2D per-pixel features $f_{i}^{2D}$ from pixel-aligned VLMs are valid for every 2D pixel, ensuring that the denominator is never zero.

\subsection{Geometric Guided Multi-View Fusion}
\label{sec3.3:refine3D}
Given the improved multi-view 2D features $F^{2D^{'}}=f^{2D^{'}}_{1}, ..., f^{2D^{'}}_{N}\in\mathbb{R}^{H \times W \times C}$ and the input 3D scene points $P\in\mathbb{R}^{M \times 3}$, our next step is to associate each 3D point with its corresponding open-vocabulary features. In addition to calculating 3D point features $F^{3D^{'}}$ by averaging multi-view pixel features $F^{2D^{'}}$ through point-pixel alignment~\cite{peng2023openscene}, we further refine these features with geometric priors from 3D perspectives. We assume that points with similar geometry in 3D space should exhibit similar features. Specifically, based on the normal similarity, we employ a graph-cut algorithm to generate $Q$ superpoints~\cite{felzenszwalb2004efficient, papon2013voxel} $U_{q}\in\mathbb{R}^{N_q \times 3} (q=1,...,Q)$ for $P$, where $N_q$ represents the number of points in $q$-th superpoint. Ideally, the points within each superpoint belong to the same semantic category. Thus, with geometric guided multi-view fusion, the final per-point 3D features $F^{3D^{''}}$ can be calculated by averaging the 3D features within the superpoints:
\vspace{-0.2cm}
\begin{equation}
\vspace{-0.2cm}
    F^{3D^{''}}(p_{k}) = \frac{1}{N_q}\sum_{p_j\in U_q}F^{3D^{'}}(p_j),
    \label{eq:5}
\end{equation}
where $F^{3D^{'}}(p_j)$ is the initial 3D features of point $p_{j}$, $p_{k} \in U_{q}$ is the $k$-th point in superpoint $U_{q}$, and $F^{3D^{''}}(p_{k})$ is the final features of point $p_{k}$.  

Finally, with refined per-point features and text features obtained by encoding arbitrary open set text prompts with CLIP text encoder, we can compute the cosine similarities between them. During inference, each point is then assigned to the text with the highest similarity.

 \begin{table*}[t]
	\centering
	\begin{tabular}{c|c|cc|cc|cc}
         \hline
		 Methods (time order) &  Replica & \multicolumn{2}{c|}{Matterport80} & \multicolumn{2}{c|}{Matterport160} & \multicolumn{2}{c}{ScanNet200} \\
        \hline
	         & mIoU  & mIoU & mAcc & mIoU & mAcc & mIoU & mAcc \\
	\hline
          MinkowskiNet~\citep{choy20194d} [CVPR'19]  &- & - &- & - &- & 25.0 & - \\
          LGround~\citep{rozenberszki2022language} [ECCV'22]  &- & - &- & - &- & 27.2 & - \\
          PTv3~\citep{wu2024point} [CVPR'24]  &- & - &- & - &- & 36.0 & - \\
        \hline
          PLA~\citep{ding2023pla} [CVPR'23]  &- & - &- & - &- &1.8 & - \\
          OpenScene-2D~\citep{peng2023openscene} [CVPR'23] &18.0 & 21.6 &27.5 & 14.2 & 20.2 &12.7 & 25.5\\
          OpenScene-3D~\citep{peng2023openscene} [CVPR'23] &11.1 & 18.1 &27.3 &8.9 &14.1 &6.3 & 12.2\\
          OpenScene-2D3D~\citep{peng2023openscene} [CVPR'23]  &14.9  & 21.1& 33.3 & 13.1 &22.7 &11.6 & 21.6\\
          RegionPLC~\citep{yang2024regionplc} [CVPR'24]  &- & - &- & - &- &6.5 & - \\
          OV3D~\citep{jiang2024open} [CVPR'24]  &- & - &- & - &- &8.7 & - \\
          GGSD~\citep{wang2024open} [ECCV'24]  &- & 11.9 &16.2 & 6.3 &9.3 & - &-\\
          DMA (OpenSeg)~\citep{li2024dense} [ECCV'24]  &- & 19.7 &- &9.4 &- & 7.6 &14.6 \\
          DMA (FC-CLIP)~\citep{li2024dense} [ECCV'24]  &-  & 20.1 &- &9.8 &- & 7.9 &15.2\\
        \hline
          MVOV3D (Ours)  &\textbf{20.1}  & \textbf{23.9} & \textbf{28.3} &\textbf{16.2} & \textbf{21.5}&\textbf{14.7} & \textbf{27.0}\\
	\hline
	\end{tabular}
        \vspace{-0.1cm}
        \caption{Performance comparison with state-of-the-art methods for open-vocabulary 3D semantic segmentation on challenge Replica, Matterport80, Matterport160, and ScanNet200 datasets.}
        \vspace{-0.3cm}
	\label{tab1}
\end{table*}

\section{Experiments}
\label{sec:experiments}

\subsection{Experiment Setup}
\paragraph{Datasets.}To demonstrate the effectiveness of our proposed approach, we conduct experiments across widely used benchmarks including ScanNet200~\citep{rozenberszki2022language}, Replica~\citep{straub2019replica}, and Matterport3D~\citep{chang2017matterport3d} for challenging open-vocabulary 3D semantic segmentation task. We also use ScanNet200 for open-vocabulary 3D instance segmentation task. ScanNet200 provides 1,613 (1201 for training and 312 for testing) indoor 3D scenes from 2.5 million RGB-D video views. ScanNet200 contains more challenging 200 semantic labels, which is well-suited for evaluating real-world open-vocabulary scenarios with a long-tail distribution. We further experiment with Replica dataset (51 categories) on office0, office1, office2, office3, office4, room0, room1, room2 scenes. Matterport3D comprises 90 highly detailed scenes of 194k RGB-D images. Besides 21 semantic labels from original Matterport3D benchmark, we also use 80, 160 most frequent classes of the NYU label set provided with the benchmark for more challenging open-vocabulary evaluation~\citep{peng2023openscene, wang2024open, li2024dense}. The mean Intersection-of-Union (mIoU) and mean Accuracy (mAcc) are employed as the evaluation metrics.


\vspace{-0.3cm}
\paragraph{Implementation details.}Our framework employs OpenSeg~\cite{ghiasi2022scaling} as pixel-aligned VLM, Grounding-DINO~\cite{liu2023grounding}, and SAM~\cite{kirillov2023segment} as VFM to segment 2D masks for obtaining image regions, and RAM++~\cite{huang2023open} as image caption models to generate text proposal.

\begin{table}[t]
	\centering
        \begin{small}
	\begin{tabular}{c|ccc}
         \hline
		 Method (time order)& AP & AP$_{50}$ & AP$_{25}$ \\
	\hline
         ISBNet~\cite{ngo2023isbnet} [CVPR'23] & 24.5 & 32.7 & 37.6  \\
         Mask3D~\cite{schult2023mask3d} [ICRA'23] & 26.9 & 36.2 & 41.4  \\
         \hline
         OpenScene~\cite{peng2023openscene} + DBScan~\cite{ester1996density} & 2.8 & 7.8 &  18.6 \\
         OpenScene~\cite{peng2023openscene} + Mask3D~\cite{schult2023mask3d} & 11.7 & 15.2 & 17.8  \\
         OVIR-3D~\cite{lu2023ovir} [CoRL'23]& 13.0 & 24.9 &  32.3 \\
         OpenMask3D~\cite{takmaz2023openmask3d} [NeurIPS'23]& 15.4 & 19.9 &  23.1 \\
         Open3DIS~\cite{nguyen2024open3dis} [CVPR'24]& 18.2 & 26.1 &  31.4  \\
         UniSeg3D~\cite{xu2024unified} [NeurIPS'24]& 19.7 & - &  - \\
         SOLE~\cite{lee2024segment} [ICLR'25]& 20.1 & 28.1 &  33.6  \\
         MVOV3D + SOLE (Ours)& \textbf{20.7} & \textbf{29.7} &  \textbf{35.6}  \\
	\hline
	\end{tabular}
        \end{small}
        \vspace{-0.1cm}
        \caption{Comparison of open-vocabulary 3D instance segmentation on ScanNet200.}
        \vspace{-0.4cm}
	\label{tab:scannet200_instance_seg}
\end{table}

\subsection{Main Results}
In this section, we compare our MVOV3D with other state-of-the-arts within open-world scenarios for open-vocabulary 3D semantic segmentation and instance segmentation. We test all approaches on the ScanNet200 validation set, Matterport3D test part, and Replica test split.

\vspace{-0.3cm}
\paragraph{Zero-Shot 3D Semantic Segmentation Comparison.}To evaluate the open-vocabulary capability of the existing methods, we conduct experiments in challenging settings on Replica, Matterport80, Matterport160, and ScanNet200 datasets, each having a large vocabulary of 51, 80, 160, and 200 categories, respectively. We present the quantitative results on these datasets in Table \ref{tab1}. Our framework reduces the performance gap between open-vocabulary methods with zero-shot inference and fully-supervised approaches, such as MinkowskiNet~\cite{choy20194d} and PTv3~\cite{wu2024point}. Notably, our MVOV3D demonstrates superior performance in open-vocabulary 3D semantic segmentation with 20.1\% mIoU on Replica51, 23.9\% mIoU on Matterport80, 16.2\% mIoU on Matterport160, and 14.7\% mIoU on ScanNet200, respectively. Remarkably, our MVOV3D outperforms all previous approaches, including the recent 3D-2D distillation methods DMA~\cite{li2024dense} and GGSD~\cite{wang2024open}, as well as 3D-1D contrastive methods RegionPLC~\cite{yang2024regionplc} and OV3D~\cite{jiang2024open}, by a significant margin. For example, our MVOV3D surpasses RegionPLC of 8.2\% mIoU and OV3D of 6.0\% mIoU on ScanNet200, respectively. Additionally, MVOV3D shows +3.8\%, +6.4\%, and +6.8\% improvement over DMA in mIoU on Matterport80, Matterport160, and ScanNet200.

These results unveil the potential upbound of current 3D open-vocabulary approaches in real-world settings, \textbf{revealing that training a 3D network on a specific dataset—whether using explicit point-text contrastive learning or implicit point-pixel distillation—fails to achieve robust open-vocabulary capabilities.} The scarcity and limited size of existing 3D datasets constrain their potential to train a 3D network effectively for open-world scenarios.

Furthermore, we observe that OpenScene-2D3D~\cite{peng2023openscene}, which combines both 2D multi-view fusion features and 3D distilled features for open-vocabulary tasks, performs even worse than OpenScene-2D, which relies solely on 2D multi-view features. This finding contrasts with prior works \cite{peng2023openscene, wang2024open, li2024dense, jiang2024open}, which conduct experiments on standard ScanNet~\cite{dai2017scannet} and Matterport3D~\cite{chang2017matterport3d} datasets with limited vocabularies of 20 and 21 classes, and states that combined 2D and 3D features enhance robustness in open-vocabulary scenarios. However, our experiments on Matterport160 and ScanNet200, which include larger vocabularies and are better suited for open-vocabulary evaluations, indicate that 3D distilled features struggle with rare and long-tail categories. Although 3D-2D distillation methods leverage features from powerful foundation models as supervision, training on a limited 3D dataset primarily populated with common categories does not fully transfer open-vocabulary capabilities to rare classes. Consequently, while 3D distilled features perform well in frequent classes, they fail with rare categories, restricting their application in the real world. More experiments about detailed quantitative results about long-tail categories and extending our MVOV3D to 3D-2D distillation framework can be found in our supplementary materials.


\begin{figure*}[t]
  \centering
   \includegraphics[width=0.92\linewidth]{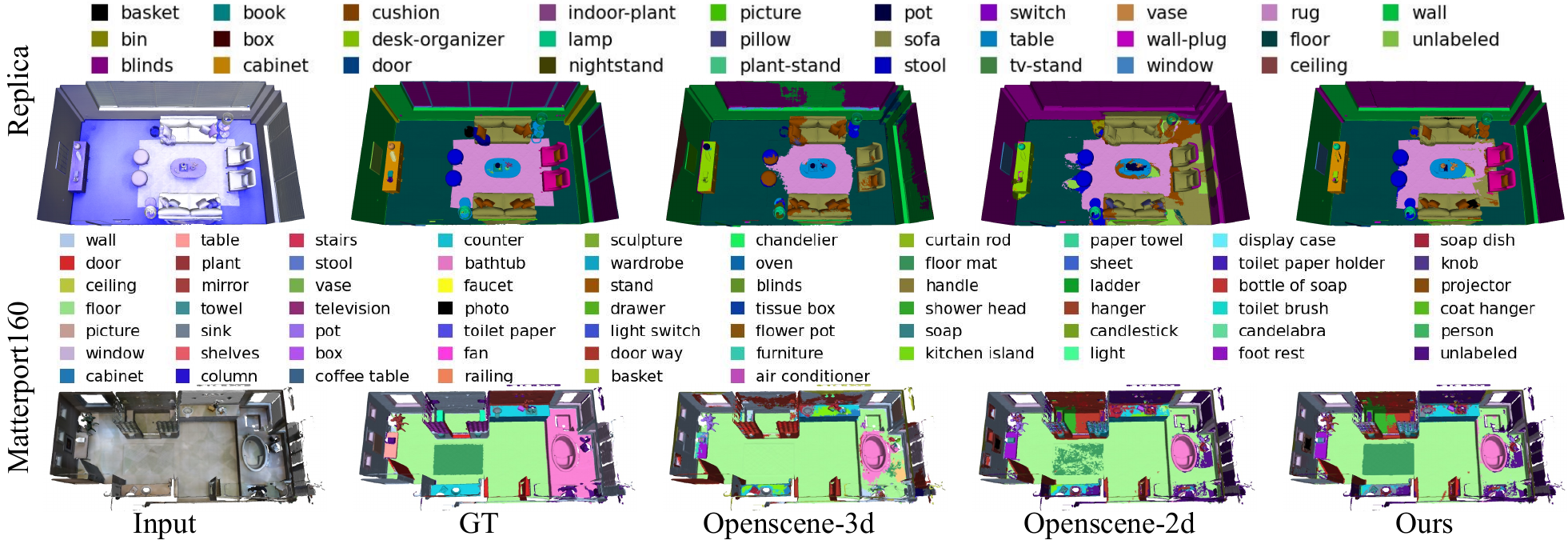}
   \vspace{-0.2cm}
   \caption{Qualitative results of zero-shot 3D semantic segmentation from our model and OpenScene.}
   \vspace{-0.2cm}
   \label{fig:figure5_visualization}
\end{figure*}


\vspace{-0.3cm}
\paragraph{Extending MVOV3D to Zero-Shot 3D Instance Segmentation.}To further evaluate the open-world understanding capabilities of our method, we extend our MVOV3D to zero-shot 3D instance segmentation task on the ScanNet200 benchmark. We implement it by using our per-point features to substitute the initial per-point CLIP features of SOLE~\cite{lee2024segment}. Predicted instance masks and mask features are obtained by their instance query through their cross-modality decoder. According to Table \ref{tab:scannet200_instance_seg}, combining our MVOV3D with SOLE yields notable 20.7\% AP, which boosts over SOLE +0.6\% and surpasses recent proposed UniSeg~\cite{xu2024unified} +1.0\% AP.

\vspace{-0.3cm}
\paragraph{Qualitative Results.}To illustrate the effectiveness of our MVOV3D in understanding open-vocabulary 3D scenes, we present the visualization of zero-shot semantic segmentation in Fig. \ref{fig:figure5_visualization}. Our MVOV3D successfully segments categories such as cushion, vase, and floor mat that are rare in the benchmarks, while other methods fail to do so. These qualitative results demonstrates the excellent open vocabulary capability of our model.


\begin{table}[t]
	\centering
	\begin{tabular}{c|c|c}
         \hline
		 Method  & Matterport160 & Replica \\

	\hline
         Our full model  & \textbf{16.2} & \textbf{20.1} \\
         \hline
         w/o 1D features $f^{1D}_{T_{i,s,select}}$ & 15.2 & 18.8 \\
         w/o 2D features $f_{R_{i,s}}^{2D}$ & 16.0 & 19.2 \\
         w/o 3D priors $U_{q}$ & 15.9 & 19.9 \\
	\hline
	\end{tabular}
        \vspace{-0.1cm}
        \caption{Ablation studies of each component of our MVOV3D method. We report the mIoU of different settings on Matterport160 and Replica.}
        \vspace{-0.2cm}
	\label{tab5:ablation}
\end{table}

\begin{table}[t]
	\centering
	\begin{tabular}{cc|cc}
         \hline
		 VFM  & mIoU & ICM & mIoU \\
	\hline
         SAM~\cite{kirillov2023segment}  & 14.9 & DeCap~\cite{lidecap} & 16.0\\
         GD~\cite{liu2023grounding} & 15.9 & BLIP~\cite{li2022blip} & 16.0\\
         SAM \& GD & \textbf{16.2}  & RAM++\cite{huang2023open} &\textbf{16.2} \\
	\hline
	\end{tabular}
        \vspace{-0.1cm}
        \caption{Effects of various VFM and image caption models.}
        \vspace{-0.4cm}
	\label{tab6:VFM_image_caption}
\end{table}

\subsection{Ablation Study}

\paragraph{Component Analysis.} As shown in Table \ref{tab5:ablation}, we conduct ablation studies on Matterport160 and Replica datasets to assess the individual contribution of each component in our MVOV3D. The complete model achieves the highest performance with 16.2\% mIoU on Matterport160 and 20.1\% mIoU on Replica. Removing 1D text features $f^{1D}_{T_{i,s,select}}$, which are extracted from confident text descriptions, leads to a noticeable decrease in performance, dropping the mIoU to 15.2\% on Matterport160 and 18.8\% on Replica. This emphasizes the rich vocabularies of image caption models in capturing diverse concepts in 3D scenes. Without incorporating 2D features $f_{R_{i,s}}^{2D}$, the performance degrades by -0.2\% and -0.9\%, which indicates that the precise 2D masks provided by VFM are beneficial for outcoming occlusion issues. Both additional 2D and 1D features help mitigate inherent noise in VLM before multi-view fusion. Finally, discarding 3D guided multi-view fusion, which refines per-point features by considering geometry priors, also results in performance decrements. This highlights that 3D geometric priors $U_{q}$ help alleviate inherent noise in point-level features after fusion. More experiment about parameter analysis can be found in our supplementary materials.

\vspace{-0.3cm}
\paragraph{VFM for 2D Masks \& ICM for 1D Text Proposal.} In Table \ref{tab6:VFM_image_caption}, we experiment with various VFM to obtain image regions for extracting image region features $f_{R_{i,s}}^{2D}$ and different image caption models to describe text for extracting text features $f^{1D}_{T_{i,s,select}}$ on Matterport160. For 2D refinement, we compare the effectiveness of SAM~\cite{kirillov2023segment} and Grounding-DINO (GD)~\cite{liu2023grounding} as standalone models and in combination with SAM \& GD for obtaining image regions and extracting corresponding 2D region features $f_{R_{i,s}}^{2D}$. The combined SAM \& GD configuration achieves the highest mIoU of 16.2\%. For 1D refinement, we explore several image caption models such as DeCap~\cite{lidecap}, BLIP~\cite{li2022blip}, and RAM++~\cite{huang2023open} for generating textual descriptions and extract text features $f^{1D}_{T_{i,s,select}}$. All models perform comparably well, with RAM++ yielding a slight advantage. These findings indicate that using advanced VFMs and image caption models can substantially enhance open-vocabulary 3D scene understanding. As foundation models continue to evolve, our method’s performance potential will be further amplified, paving the way for more robust and scalable solutions in open-world 3D understanding.

\section{Conclusion}
\label{sec:conclusion}
We analyze the limitations of current 3D-2D distillation and 3D-1D contrastive methods in training 3D networks on relatively small 3D datasets. Based on this, we propose MVOV3D, a framework that reduces the inherent noises in multi-view features by leveraging multiple foundation models. We abandon the 3D networks and focus on preserving the open-vocabulary capabilities of pre-trained VLMs. We hope MVOV3D establishes a strong baseline for 3D open-vocabulary understanding for the community. Future work could explore the integration of highly generalizable foundation models from 1D and 2D domains into 3D with 3D priors or the development of a 3D network with robust open-vocabulary capabilities.
\vspace{-0.4cm}
\paragraph{Limitations.} Our method relies on the quality of various foundation models, including 2D masks generated by VFM, text descriptions produced by image captioning models, and 2D open-vocabulary features extracted via VLM. With the advancement of foundation models, the potential performance of our method will be further unleashed.
\clearpage
\setcounter{page}{1}
\maketitlesupplementary

\setcounter{section}{0}
\renewcommand{\thesection}{\Alph{section}}


\begin{table}[t]
	\centering
	\begin{tabular}{c|c|c|c}
         \hline
		 Method & head & common & tail \\
	\hline
         OpenScene-3D &16.5/25.0 & 10.9/21.9 &6.0/12.5 \\
         OpenScene-2D &\textbf{21.5/30.8} & 15.6/35.4 &17.4/32.0 \\
         MVOV3D (Ours) &20.8/29.7 & \textbf{21.1/45.7} &\textbf{18.5/34.1} \\
        \hline
         OpenScene-3D &24.1/32.9 & 5.8/7.8 &1.1/4.5 \\
         OpenScene-2D &25.7/27.8 & 15.3/20.6 &9.7/18.6 \\
         MVOV3D (Ours) &\textbf{29.0/31.4} & \textbf{17.8/25.5} &\textbf{10.8/22.1} \\
        \hline
	\end{tabular}
        \vspace{-0.2cm}
        \caption{Detailed results (mIoU/mAcc) about long-tail categories on Replica (top) and Matterport160 (bottom).}
        \vspace{-0.2cm}
	\label{tab: Replica_Matterport160_tail}
\end{table}

\section{Results about Long-Tail Categories.}
We further provide detailed quantitative results of open-vocabulary 3d semantic segmentation on Replica and Matterport160 datasets in Table \ref{tab: Replica_Matterport160_tail}. Both labels are categorized into three subsets—head (17 and 53 categories), common (17 and 54 categories), and tail (17 and 53 categories)—based on the frequency of labeled points in the datasets. The results indicate that distilling 3D network (OpenScene-3D) on relative small 3D datasets can significantly \textbf{\textit{impair}} performance in rare classes (1.1\% mIoU of tail classes on Matterport160). In contrast, our MVOV3D \textit{maintains \textbf{better}} open-vocabulary capabilities for long-tail categories (18.5\% and 10.8\% mIoU of tail classes on Replica and Matterport 160).

\begin{table}[t]
	\centering
	\begin{tabular}{c|c|c}
         \hline
		 Methods  & ScanNet200 & Matterport160 \\
	\hline
         OpenScene-3D  & 6.3 &8.2 \\
        \hline
          MVOV3D-3D (Ours)  & \textbf{6.9} (+0.6)  &\textbf{9.2} (+1.0)\\
	\hline
	\end{tabular}
        \vspace{-0.2cm}
        \caption{Comparisons between MVOV3D-3D and OpenScene-3D on 3D-2D distillation.}
        \vspace{-0.7cm}
	\label{tab3:distill}
\end{table}

\vspace{-0.3cm}
\section{Extending MVOV3D to 3D-2D Distillation.}
Our MVOV3D, originally designed to reduce inherent noise in 2D multi-view features without requiring additional training, also enhances the 3D distillation process. When some tasks provide only 3D point clouds or meshes, our refined features can be used to distill a 3D network for improved performance. As presented in Table \ref{tab3:distill}, our MVOV3D achieves a mIoU increase of +0.6\% on ScanNet200 and +1.0\% on Matterport160 compared with OpenScene-3D, respectively. This demonstrates the effectiveness of our framework on alleviating noise in 2D multi-view features, which in turn can benefit the 3D-2D distillation process.

\vspace{-0.3cm}
\section{More Visualization.}
\begin{figure}[t!]
  \centering
   \includegraphics[width=0.85\linewidth]{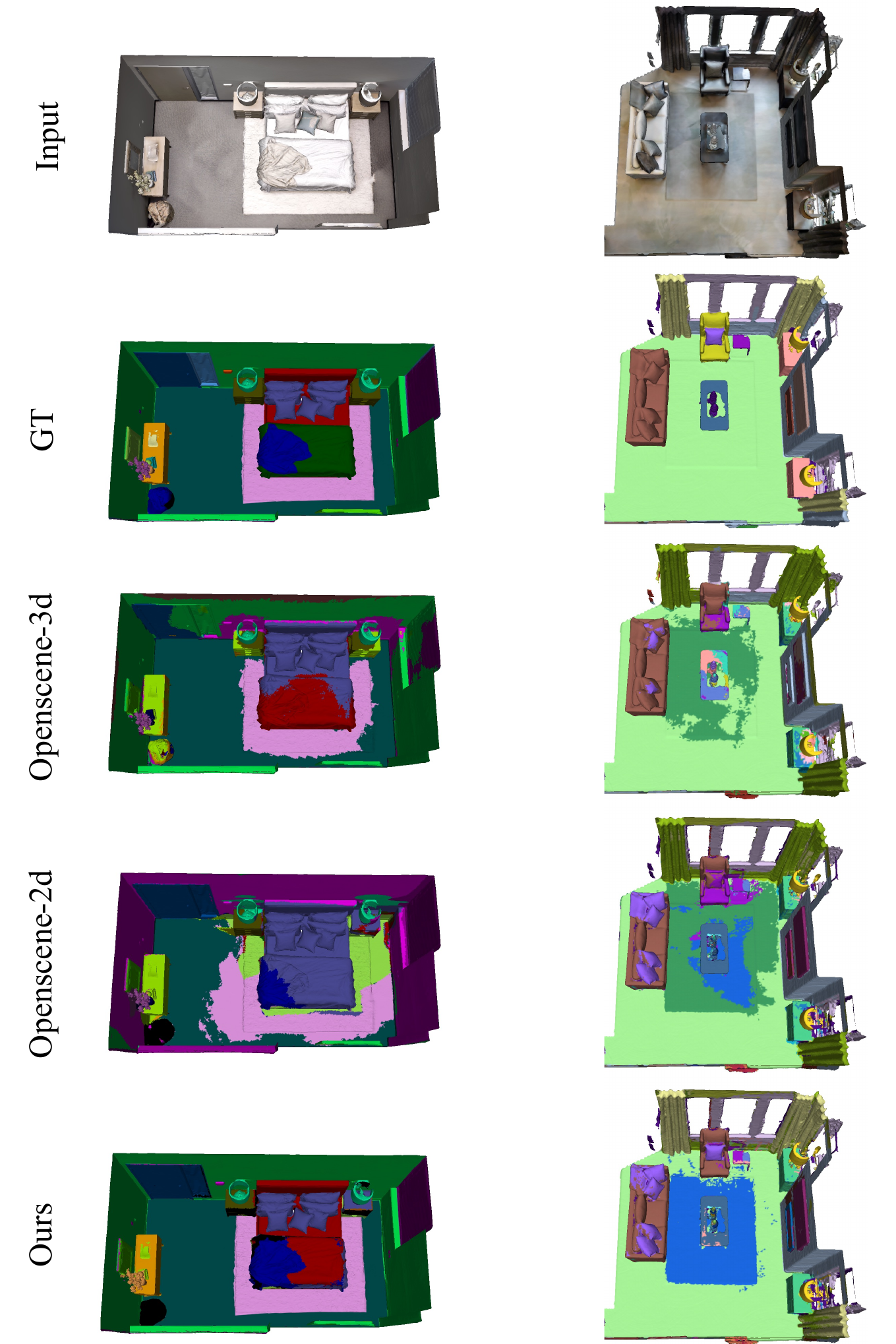}
   \caption{Qualitative results of zero-shot 3D semantic segmentation from our model and OpenScene.}
   \label{fig:supplement_visualization}
\end{figure}

Additional qualitative results are shown in Fig. \ref{fig:supplement_visualization}

\begin{figure}[t]
  \centering
   \includegraphics[width=0.82\linewidth]{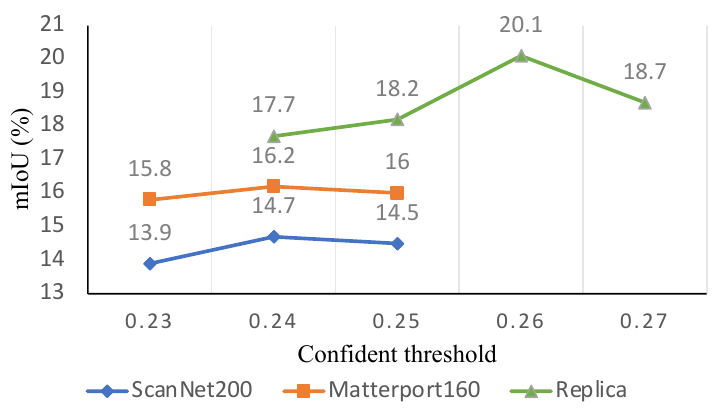}
   \vspace{-0.3cm}
   \caption{Parameter experiments about confident threshold $\delta$.}
   \label{fig:figure7_parameter_analysis}
\end{figure}

\section{Parameter Analysis.}
We also conduct experiments with different confident threshold $\delta$ on Replica, Matterport3D and ScanNet200 datasets. As illustrated in Fig. \ref{fig:figure7_parameter_analysis}, when $\delta$ is set to 0.24 on Matterport3D and ScanNet200, and 0.26 on Replica, leading to better performance.


\clearpage
{
    \small
    \bibliographystyle{ieeenat_fullname}
    \bibliography{main}
}

\end{document}